\documentclass[runningheads]{llncs}
\usepackage{graphicx}
\usepackage{amssymb}
\usepackage{amsmath}
\usepackage{textcomp}
\usepackage{xcolor}
\usepackage{diagbox} 
\usepackage{algorithm}

\usepackage{multirow}
\begin{document}
\title{A Model-Agnostic SAT-based approach for Symbolic Explanation Enumeration \\(Preprint version)}

\author{Ryma Boumazouza \orcidID{0000-0002-3940-8578}
 \and
Fahima Cheikh-Alili \orcidID{0000-0002-4543-625X} 
\and
Bertrand Mazure \orcidID{0000-0002-3508-123X}
 \and
Karim Tabia \orcidID{0000-0002-8632-3980}
}
\authorrunning{R. Boumazouza et al.}
%
\institute{CRIL, Univ. Artois and CNRS, F62307 Lens, France
\email{\{ryma.boumazouza,cheikh,bertrand.mazure,karim.tabia\}@univ-artois.fr}
}

\maketitle              
%
%
\begin{abstract}
We propose a generic agnostic approach allowing to generate different and complementary types of symbolic explanations. 
More precisely, we generate explanations to locally explain a single prediction by analyzing the relationship between the features and the output.
 Our approach uses a propositional encoding of the predictive model and a SAT-based setting to generate two types of symbolic explanations which are \textit{Sufficient Reasons} and \textit{Counterfactuals}. 
 The experimental results on imagee classification task show the feasibility of the proposed approach and its effectiveness in providing \textit{Sufficient Reasons} and \textit{Counterfactual} explanations.

\end{abstract}
\section{Introduction}
Modern machine learning (ML) and deep learning methods are nowadays widely used in many sensitive fields and industries. However, despite the good predictive performance of the ML models, there are critical applications that fundamentally require trust like applications applied to medicine, driverless	cars and law enforcement. Therefore, it becomes increasingly important to explain the behavior of those models and their output to enhance trust in the model predictions and their adoption in real world applications. This leads to a rapid growth in attention to eXplainable AI (XAI). The XAI methods can be grouped into pre-model (ante-hoc), in-model, and post-model (post-hoc) methods. In the latter, we identify two types of explanations: (1) symbolic (knowledge-driven) methods that are based on logical representations used for explanation (e.g. \cite{shih2018symbolic},\cite{ignatiev2019relating}), verification and diagnostic purposes (e.g. \cite{reiter1987theory}, \cite{rymon1994se}, \cite{ignatiev2019relating}), and (2) numerical feature-based methods that provide insights into how much each feature contributed to that outcome (e.g. SHAP\cite{lundberg2017unified}, LIME\cite{ribeiro2016should}). The main limitation to the existing explainability methods based on symbolic representations is the fact that they are generally intended to specific models and cannot be applied to any model (non agnostic). In the other hand, feature-based methods such as LIME\cite{ribeiro2016should} and SHAP\cite{lundberg2017unified} try to assess the amount of contribution of features into the predictions but fail at answering certain questions such as {\it What are the feature values which are sufficient in order to trigger the prediction whatever the values of the other variables ? } or {\it Which values are sufficient to change in the instance $x$ to have a different prediction ?} \\To address the problem of answering this type of fundamental questions, we propose in this paper an approach to provide "different" and "complementary" types of symbolic explanations: the {\bf \textit{Sufficient Reasons} (\textbf{$SRx$} for short)} and the {\bf \textit{Counterfactuals} (\textbf{$CFx$} for short)}.
The main advantage of our approach is the fact of being model-agnostic, where we try to learn accurate yet simple models to emulate the given black-box.
In the other hand, the approach is based on rigorous and well-known Boolean satisfiability concepts that allow us to exploit the availability of efficient SAT solvers. Accordingly, our approach is declarative and does not require the implementation of specific algorithms. We evaluate the feasibility and efficiency of our approach on image classification task.

\section{Preliminaries and notations}
We first introduce the necessary notations and recall some definitions used in the remainder of this paper. 
For the sake of simplicity, we will limit the presentation to binary classifiers with binary features. We also focus only on negative predictions where the outcome is $0$. As for explaining positive predictions where the outcome is $1$, the approach applies similarly as discussed in the Conclusion.
\begin{definition}{\textbf{(Binary Classifier)}} A Binary Classifier is defined by two sets of variables: A feature space $X$= \{$X_1$,...,$X_n$\} where $|X|=n$, and a binary class variable denoted $Y$. Both the features and the class variable take values $in$ \{0,1\}.
\end{definition}
A decision function describes the classifier's behavior independently from the way it is implemented. We define it as a function $f:X \rightarrow Y$ mapping each instantiation  $x$ of $X$ to $y$=$f(x)$. A data instance $x$ is the feature vector associated with an instance of interest whose prediction from the ML model is to be explained. We use interchangeably in this paper $f$ to designate the classifier and its decision function. Let us now define the representation framework we use.
\begin{definition}{\textbf{(SAT : The Boolean Satisfiability problem)}}
Usually called SAT, the Boolean satisfiability problem is the decision problem, which, given a propositional logic formula, determines whether there is an assignment of propositional variables that makes the formula true.
\end{definition}
The logic formulas are built from propositional variables and Boolean connectors "AND" ($\wedge$), "OR" ($\vee$), "NOT" ($\neg$). A formula is satisfiable if there is an assignment of all variables that makes it true. It is said inconsistent or unsatisfiable otherwise.
A complete assignment of variables making a formula true is called a model while a complete assignment making it false is called a countermodel. 
 \begin{definition}{\textbf{(CNF (Clausal Normal Form))}}
is a set of clauses seen as a conjunction. A clause is a formula composed of a disjunction of literals. A literal is either a Boolean variable $p$ or its negation $\neg p$. A quantifier-free formula is built from atomic formulae using conjunction $\wedge$, disjunction $\vee$, and negation $\neg$. An interpretation $\mu$ assigns values from \{0, 1\} to every Boolean variable. Let $\Sigma$ be a CNF formula, $\mu$ satisfies  $\Sigma$ iff $\mu$ satisfies all clauses of $\Sigma$.
 \end{definition}
  Thanks to the achievements that the SAT field had known in the recent years, the modern SAT solvers\footnote{A SAT solver is a program for establishing the satisfiability of Boolean formulas encoded in conjunctive normal form.} have gained in performance and efficiency where they can handle now problems with several million clauses and variables.
  Recall that we use a SAT oracle to generate our formal symbolic explanations. We encode the explanation generation problem as two common problems related to SAT-solving which are enumerating {\it minimal reasons why a formula is inconsistent} and {\it minimal changes to a formula} in order to restore its consistency. Indeed, in the case of an unsatisfiable CNF, we can analyze the inconsistency by enumerating sets of clauses causing the inconsistency (called  Minimal Unsatisfiable Subsets (MUS)), and other sets of clauses allowing to restore its consistency (called Minimal Correction Subsets (MCS)). The enumeration of MUS/MCS are well-known problems dealt with in many areas such as knowledge-base reparation. Several approaches and tools have been proposed in the SAT community  for their generation (e.g \cite{liffiton2008algorithms}, \cite{gregoire2007boosting}).
 

\section{Related Works}
Explaining machine learning systems has been a hot research topic recently.
There has been hundreds of papers on ML explainability but we will be focusing on the ones closely related to our work.
In the context of model-agnostic explainers where  the learning
function of the input model and its parameters are not known (black-box), we can cite some post-hoc explanations methods such as: LIME (Local Interpretable Model-agnostic Explanations) \cite{ribeiro2016should} which explain black-box classifiers by training an interpretable model $g$ on samples randomly generated in the vicinity of the data instance. We follow an approach similar to LIME, the difference is that we encode our surrogate model into a CNF to generate symbolic explanations. 
The authors in \cite{ribeiro2018anchors} proposed a High-precision model agnostic explanations called ANCHOR. It is based on computing a decision rule linking the feature space to a certain outcome, and consider it as an anchor covering similar instances. Something similar is done in SHAP (SHapley Additive exPlanations) \cite{lundberg2017unified} that provides explanations in the form of the game theoretically optimal called Shapley values. Due to its computational complexity, other model-specific versions have been proposed for linear models and deep neural networks (resp LinearSHAP and DeepSHAP) in \cite{lundberg2017unified}.
The main difference with this rule sets/feature-based explanation methods is that we propose symbolic explanations using a formal method based on the logical representation of a model, and we use a substitution approach to make the model method agnostic. 

Recently, some authors propose symbolic and logic-based XAI approaches that can be used for different purposes \cite{darwiche2020}. We can distinguish the compilation-based approaches where Boolean decision functions of classifiers are compiled into some symbolic forms. For instance, in \cite{chan2012reasoning,shih2018symbolic} the authors showed how to compile the decision functions of naive Bayes classifiers into a symbolic representation, known as Ordered Decision Diagrams (ODDs).
{We proposed in a previous work \cite{boumazouza2020symbolic} an approach designed to equip such symbolic approaches \cite{shih2018symbolic} with a module for counterfactual explainability.
There are some ML models whose direct encoding into CNF is possible. For instance, the authors in \cite{narodytska2018verifying} proposed a CNF encoding for Binarized Neural Networks (BNNs) for verification purposes. In \cite{shi2020tractable}, the authors propose a compilation algorithm of BNNs into tractable representations such as Ordered Binary Decision Diagrams (OBDDs) and Sentential Decision Diagrams (SDDs).  The authors in \cite{shih2019compiling} proposed  algorithms for compiling Naive and Latent-Tree Bayesian network classifiers into decision graphs. In \cite{KR2020-86}, the authors dealt with  a set of explanation queries and their computational complexity once classifiers are represented with compiled representations. 
However, the compilation-based approaches are hardly applicable to large sized models, and remain strongly dependent on the type of classifier to explain (non agnostic). Our approach can use those compilation algorithms to represent the whole classifier when the encoding remains tractable, but in addition, we propose a local
approximation of the original model using a surrogate model built on the neighborhood of the instance at hand. 

Recent works already address the enumeration of symbolic
explanations using linear programming, mixed-integer programming, or SMT solvers.  Authors in \cite{ignatiev2019relating,ignatiev2019abduction,marques2020explaining} deal with some explanations referred to as abductive explanations (AXp) and contrastive explanations (CXp) using SMT oracles and cover the enumeration of so-called generalized
explanations. The work presented in \cite{ignatiev2020contrastive} is about duality of abductive and contrastive explanations and their enumeration. In \cite{ignatiev2021sat}, the authors explain the prediction of  decision list classifiers using a SAT-based approach. Explaining random forests and decision trees is dealt with for instance in \cite{KR2020-86} and \cite{DBLP:journals/corr/abs-2012-11067,DBLP:journals/corr/abs-2010-11034} respectively.
The main difference with our work is that we propose a surrogate-based approach where we directly relate explanations with the notions of MCS and MUS in a Partial MaxSAT setting. This allows obtaining explanations thanks to the use of off-the-shelf tools for enumerating MCS and MUS.


\section{Overview of the proposed approach }
The objective of our approach is explaining the prediction made by a classifier for a given input data instance $x$. 
It associates a logical representation that is almost equivalent to the decision function of the model to explain. 
Figure \ref{fig1} represents an overview of the proposed approach. 
\begin{figure}[h!]
\centering
  \includegraphics[width=1\linewidth,height=6cm]{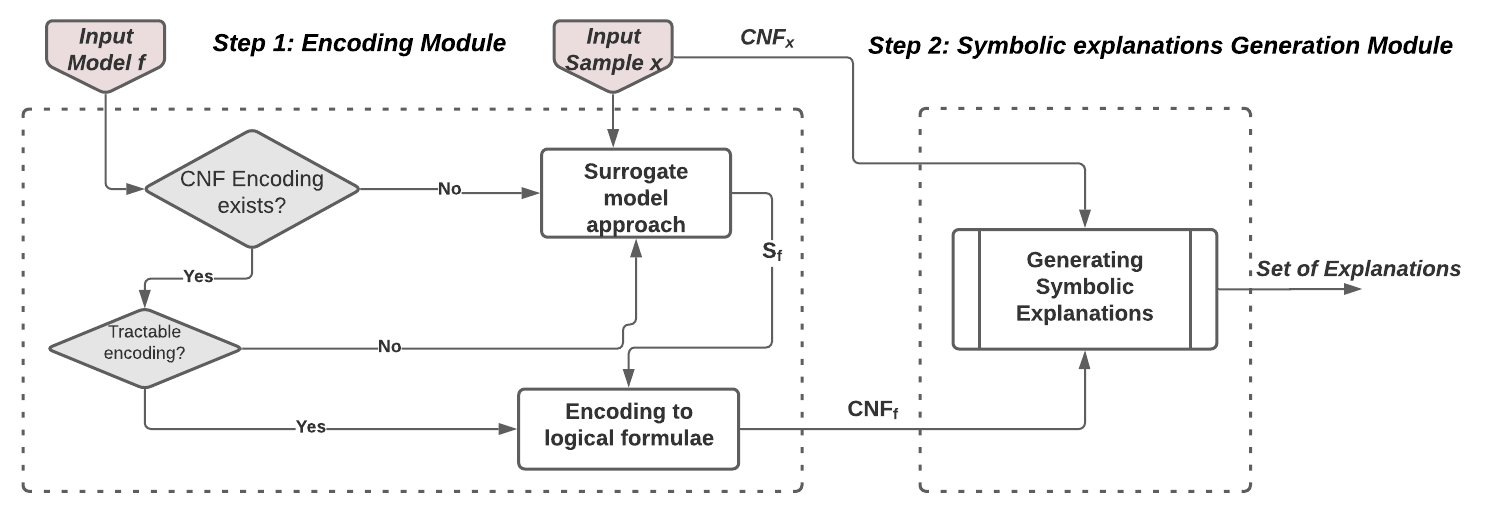} 
\caption{A global overview of the proposed approach}
\label{fig1}
\end{figure}

Given a predictive model $f$, our approach proceeds as follows:
\begin{itemize} 
\item {\bf Step 1 (CNF encoding of the classifier)}: the goal is to encode $f$ into an {\it equivalent} symbolic representation $\Sigma_f$. The generation of symbolic explanations in the next step is done using $\Sigma_f$. The encoding is done either by the means of model encoding algorithms if they are available and the encoding remains tractable, or using a surrogate modeling approach as described in Section \ref{encodingsec}. 
\item {\bf Step 2 (Explanation problem modeling)}: This step presented in Section \ref{generationsec} comes down to enumerating two types of symbolic explanations: $SRx$ and $CFx$. This task is formulated as a partial maximum satisfiability problem Partial Max-SAT\cite{biere2009handbook} using the CNF encoding $\Sigma_f$ of $f$ and $\Sigma_x$ of the input instance $x$. 
The symbolic explanations respectively correspond to Minimal Unsatisfiable Subsets(MUS) and Minimal Correction Subsets(MCS) of $\Sigma_f \cup \Sigma_x$ in the SAT terminology.
\end{itemize} 
The SAT solving is already used for providing some forms of symbolic explanations for some specific ML \cite{boumazouza2020symbolic,ignatiev2021sat,izza2021explaining}. The novelty in our approach is the fact of being model-agnostic and succeeding in generating formal explanations based on rigorous concepts.
\section{ CNF encoding of classifiers}\label{encodingsec}  
The starting point of our approach is the encoding of the input ML model $f$ into a logical representation (CNF). This step is necessary in order to use a SAT oracle for the enumeration of the symbolic explanations.
Mainly, two cases are considered: (1) an encoding of classifier $f$ into an equivalent logical representation exists, in which case we can use it (e.g. Binarized Neural Networks (BNNs) \cite{narodytska2018verifying}, Naive and Latent-Tree Bayesian network \cite{shih2019compiling}). (2) We consider the classifier $f$ as a black-box and we use a surrogate model approach to approximate it in the vicinity of the instance to explain $x$ (Agnostic option). The surrogate models are used to explain individual predictions of black-box ML models.

\noindent We set the focus on the agnostic option in this paper. This latter is applied when no direct CNF encoding exists for $f$ or if the encoding is intractable.\\

\paragraph{\bf  Surrogate model encoding into CNF}   
The approach proposed uses a surrogate model mainly characterized by its faithfulness to the initial model $f$ (ensures same predictions) and its tractable logical representation (CNF). 
To ensure the local faithfulness to $f$, we use a surrogate model $f_S$ trained on data instances in the vicinity of the data instance $x$ whose prediction from the model needs to be explained. We construct the vicinity of $x$ noted $V(x,r)$ by sampling new data instances within a radius $r$ of $x$ if the dataset is available\footnote{Otherwise, we can draw new perturbed samples around $x$}.

A model that can guarantee a good trade-offs between faithfulness and giving a tractable CNF encoding is the one of random forests \cite{ho1995random}. As shown in our experimental study, the random forest accuracy reflects a good level of faithfulness and its CNF encoding size remains tractable.
The CNF encoding $f_S$ of a classifier $f$ should guarantee the equivalence of the two representations stated as follows :
\begin{definition}{\textbf{(Equivalence of a classifier and its CNF encoding)}}
A binary classifier $f$ (resp. $f_S$) can be equivalently encoded as a CNF $\Sigma_f$ (resp. $\Sigma_{f_S}$) s.t.
$f(x)$=$1$ (resp. $f_S(x)$=$1$) iff $x$ is a model of $\Sigma_f$ (resp. $\Sigma_{f_S}$).
\end{definition}
Namely, data instances $x$ predicted positively ($f(x)$=$1$) by the classifier are models of the CNF encoding the classifier. Similarly, data instances $x$ predicted negatively ($f(x)$=$0$) are countermodels of the CNF encoding the classifier.  

\subsection{CNF encoding of random forests}
In this work, we adopted the random forest\footnote{Random Forests are used for XAI purposes in some works such was \cite{audemard2020tractable,ChoiShihGoyankaDarwiche20,izza2021explaining}} as the surrogate model $f_S$.
Its associated CNF encoding resumes in i) encoding the decision trees individually and then ii) encoding the combination rule (which is a majority voting rule).\\
\paragraph{\bf Encode in CNF every decision tree :} Remember that all the features in our case are binary. Thus, each internal node of a decision tree $ DT_i $ represents a binary test on one of the features. The result of a test is either true or false. For the leaves of a decision tree, each one is annotated with the predicted class (namely, $0$ or $1$). 
The Boolean function encoded by a decision tree can be captured in CNF as the conjunction of the negation of paths leading to leaves labelled $0$.\\ 
\paragraph{\bf Encode in CNF the combination rule :} Let $y_i$ be a Boolean variable capturing the truth value of the CNF associated to a $DT_i$. Hence, the majority rule used in random forests to combine the predictions of $m$ decision trees can be seen as a cardinality constraint\footnote{In our case this constraint means that at least $t$ decision trees predicted the label $1$.} \cite{sinz2005towards} that can be stated as follows :
\begin{equation}
y \Leftrightarrow \sum_{i=1..m} y_i \geq t,
\end{equation}
where $t$ is a threshold (usually $t$=$\frac{m}{2}$). 
{\color{black}To form the CNF corresponding to the entire random forest, it suffices to conjuct the $m$ CNFs associated to the decisions trees, and, the CNF of the combination rule.} 

\section{Generating symbolic explanations}\label{generationsec}
This section will cover the presentation of both \textit{Sufficient Reasons} and \textit{Counterfactuals} explanations as well as the SAT-based setting we use to generate such explanations. This corresponds to {\bf Step 2} within the Fig.\ref{fig1}.
This step takes as input the CNF encoding of a classifier $\Sigma_f$ and a sample data instance $\Sigma_x$.

\subsection{A SAT-based setting for the enumeration of explanations}
We propose two complementary types of symbolic explanations: the {\it Sufficient Reasons} which are a minimal subset of the input data, that if fixed, lead to a given prediction and the {\it Counterfactuals} which are a minimal subset of the input data that we can act on to obtain a different outcome. 
The enumeration of those symbolic explanations in our approach is based on two very common concepts in SAT which are MUS and MCS (defined formally in the following). We use a variant of the SAT problem called Partial-Max SAT \cite{biere2009handbook} in order to restrict the explanations only to clauses encoding the input data $x$ and do not include clauses that encode the classifier $f$.\\
The Partial-Max SAT problem can be efficiently solved by the existing tools implementing the enumeration of MUSes and MCSes such as the tool in \cite{gregoire2018boosting}. 
It is composed of two disjoint sets of clauses 
where $\Sigma_H$  denotes the hard clauses (those that could not be relaxed) and $\Sigma_S$ denotes the soft ones (those that could be relaxed). 
Concretely in our approach, the set of hard clauses corresponds to $\Sigma_f$ and the soft clauses to $\Sigma_x$. 
The  CNF $\Sigma_x$ encoding the data instance $x$ is formed by unit clauses where each clause $\alpha \in \Sigma_x$ is composed of exactly one literal ($\forall \alpha \in \Sigma_x, |\alpha|= 1$)
and each literal representing a Boolean variable of $\Sigma_x$ corresponds to a Boolean variable $\{ X_i \in X\}$ where $X$ is the feature space of $f$.\\
Thanks to this Partial-MAX SAT setting, it is possible to both identify the subsets of $\Sigma_x$ responsible for the unsatisfiability of a given CNF $\Sigma_f$$\cup$$\Sigma_x$ (corresponding to $SRx$ of $f(x)$=$0$), and the subsets allowing to restore the consistency of $\Sigma_f$$\cup$$\Sigma_x$ (corresponding to $CFx$ allowing to change the prediction to $f(x)$=$1$).

\subsection{Sufficient Reason Explanations (SR$_x$)}
We are trying to find explanations that identify the relevant variables that could justify why the prediction is negative. This is carried out by identifying a subset of our input which causes the inconsistency of the CNF formula $\Sigma_f$$\cup$$\Sigma_x$ (recall that the prediction $f(x)$ is captured by the truth value of $\Sigma_f$$\cup$$\Sigma_x$).
The identified subsets  of the input $x$ represent \textit{Sufficient Reasons} for the prediction to be negative.
We formally define the \textit{Sufficient Reasons} explanations as follow:
\begin{definition}{\textbf{({\bf \textit{SR$_x$}} explanations)}}\label{sr}
Let $x$ be a data instance and $f(x)$=$0$ its prediction by the classifier $f$. A sufficient reason explanation $\textit{\~{x}}$ of $x$ is such that:
\begin{enumerate}
    \item\label{itemi} $\textit{\~{x}}$ $\subseteq x$ ($\textit{\~{x}}$ is a part of $x$)
    {\color{black}
    \item\label{itemii} $\forall \acute{x}$, $\textit{\~{x}}$ $\subset \acute{x}:$ $f(\acute{x})$=$f(x)$  ($\textit{\~{x}}$ suffices to trigger the prediction)}
    \item There is no partial instance $\hat{{x}} \subset \textit{\~{x}}$ satisfying \ref{itemi} and \ref{itemii} (minimality)
  
\end{enumerate}
\end{definition}
Intuitively, a \textit{sufficient reason} $\textit{\~{x}}$ is defined as the part of the data instance $x$ such that  $\textit{\~{x}}$ is minimal and causes the prediction $f(x)$=$0$. Namely, to explain the
classification it is "sufficient" to observe those features with disregard to the others.
We define now the Minimal Unsatisfiable Subsets :
\begin{definition}{(\textbf{MUS})}
A Minimal Unsatisfiable Subset (MUS) is a minimal subset $\Gamma$ of clauses of a CNF $\Sigma$ such that $\forall$ $\alpha$ $\in$ $\Gamma$, $\Gamma \backslash \{\alpha\}$ is
satisfiable. 
\end{definition}
A MUS for $\Sigma_f$$\cup$$\Sigma_x$ comes down to a subset of soft clauses, namely a part of $x$ that is causing the inconsistency, hence the prediction $f(x)$=0. 

\begin{proposition}\label{srmuc}
Let $f$ be a classifier, let $\Sigma_f$ be its CNF representation. Let also $x$ be a data instance predicted negatively ($f(x)=0$) and $\Sigma_f$$\cup$$\Sigma_x$ the corresponding Partial Max-SAT encoding.   Let $SR(x,f)$ be the set of \textit{Sufficient Reasons} of $x$ wrt. $f$.
Let MUS$(\Sigma_{f,x})$ be the set of MUses of $\Sigma_f \cup \Sigma_x$.
Then: 
\begin{equation}\label{musisrsiff}
        \forall \text{\~{x}} \subseteq x, \text{\~{x}} \in SR(x,f)   \iff \text{\~{x}} \in MUS(\Sigma_{f,x})
\end{equation}
\end{proposition}
Proposition \ref{srmuc} states that each MUS of the CNF $\Sigma_f$$\cup$$\Sigma_x$ is a \textit{Sufficient Reason} for the prediction $f(x)$=0 and vice versa. 

\subsection{Counterfactual Explanations (CF$_x$)}
We are also interested in another type of explanation which would allow us to figure out what changes can be made to the input data in order to alter the initial outcome. Let us formally define the concept of counterfactual explanation.

\begin{definition}{({\bf \textit{CF$_x$}} Explanations)}\label{ce}
Let $x$ be a complete data instance and $f(x)$ its prediction by the decision function of $f$. A \textit{counterfactual} explanation $\textit{\~{x}}$ of $x$ is such that:
\begin{enumerate}
    \item $\textit{\~{x}} \subseteq x$ ($\textit{\~{x}}$ is a part of x)
    \item $f(x[\textit{\~{x}}])$= 1-$f(x)$ (prediction inversion)
    \item There is no $\hat{{x}}$ $\subset$ $\textit{\~{x}}$ such that $f(x[\hat{{x}}])$=$f(x[\textit{\~{x}}])$ (minimality)
\end{enumerate}
\end{definition}
In definition \ref{ce}, the term $x[\textit{\~{x}}]$ denotes the data instance $x$ where variables included in $\textit{\~{x}}$ are inverted.
In our approach, \textit{Counterfactuals} are enumerated thanks to the Minimal Correction Subset enumeration \cite{gregoire2018boosting}. 

\begin{definition}{\textbf{(MCS)}}
A Minimal Correction Subset $\Psi$ of a CNF $\Sigma$ is a set of formulas $\Psi$ $\subseteq$ $\Sigma$ whose complement in $\Sigma$, i.e., $\Sigma$ $\backslash$ $\Psi$, is a maximal satisfiable subset of $\Sigma$.
\end{definition}
Following our modeling, an MCS for $\Sigma_f$$\cup$$\Sigma_x$ comes down to a subset of soft clauses denoted $\textit{\~{x}}$, namely a part of $x$ that is enough to remove (or reverse) in order to restore the consistency, hence to alter the prediction $f(x)$=$0$ to $f(x[\textit{\~{x}}])$=$1$.
\begin{proposition}\label{cemsc}
Let $f$ be the decision function of the classifier, let $\Sigma_f$ be its CNF representation. Let also $x$ be a data instance predicted negatively ($f(x)=0$) and $\Sigma_f$$\cup\Sigma_x$ the corresponding Partial Max-SAT encoding.
Let $CFx(x,f)$  be the set of counterfactuals of $x$ wrt. $f$.
Let MCS$(\Sigma_{f,x})$ the set of MCSs of $\Sigma_f\cup\Sigma_x$. 
Then:  
\begin{equation}\label{mcsiscfiff}
        \forall \textit{\~{x}} \subseteq x, \textit{\~{x}} \in CFx(x,f)   \iff \textit{\~{x}} \in MCS(\Sigma_{f,x})
\end{equation}
\end{proposition}

Proposition \ref{cemsc} states that each MCS of the CNF $\Sigma_f$$\cup$$\Sigma_x$ represents a $CF$ $\textit{\~{x}}$$\subseteq$$x$ for the prediction $f(x)$=0 and vice versa.

\section{Empirical evaluation}
\subsubsection{Experimentation set-up}
The black-box models considered are \textit{"one-vs-all"} binary neural networks (BNNs)\footnote{defined as a neural networks with binary weights and activations at run-time} trained on the widely used MNIST database \footnote{MNIST: handwritten digit databse, available at http://yann.lecun.com/exdb/mnist/}.
MNIST is composed of 70,000 images of size 28 × 28 pixels. 
We use the pytorch implementation\footnote{available at: https://github.com/itayhubara/BinaryNet.pytorch} of the Binary-Backpropagation algorithm "BinaryNets"\cite{NIPS2016_d8330f85} to train the BNN classifiers (one per digit from 0 to 9) on the binarized images (threshold $T=127$).
All experiments have been conducted on Intel Core i7-7700
(3.60GHz ×8) processors with 32Gb memory on Linux. 

\subsubsection{Results}
The surrogate model considered is a random forest (RF) classifier trained on the vicinity of the input sample using the hyper-parameters $nb\_trees=10$ and $max\_depth=24$. We try out different values for the radius $r$ but we only present the results for $r=250$ with an average of 200 neighbors around $x$ due to the limited number of pages. The black-box models (BNNs) trained to recognize the "0","2","5","6" and "8" digits are used as predictive models \footnote{results for the other digits are similar but not be reported because of space limitation}. Around 1000 to 1500 images were picked randomly from the MNIST database for the experimental study conducted on each classifier.\\

\noindent {\bf Evaluating the CNF encoding in practice :}
We are interested in evaluating the size of the CNF encoding using the setting mentioned above. 
We use the Tseitin Transformation \cite{tseitin1983complexity} to encode the propositional formulae into an equisatisfiable CNF formulae. 
The size of this latter is linear in the size of the original formulae. 
The results are presented in \textbf{Table 1}. The high accuracy of $f_S$ shows that the generated RF classifier provide interesting results in term of fidelity. The number of variables/clauses of the CNF indicates that the logical representation remains tractable and makes the logical representation easily handled by the current SAT-solvers which confirms the feasibility of the approach. \\
\begin{table}[h!]
\begin{center}
\scriptsize
       \resizebox{\textwidth}{!}{ \begin{tabular}{p{3cm}|p{1.6cm}|p{1.6cm}|p{1.6cm}|p{1.6cm}|p{1.6cm}|p{1.6cm}|}
        \hline& MNIST\_0&MNIST\_2  & MNIST\_5 & MNIST\_6& MNIST\_8  \\  
          \hline
   avg acc of RF &{98\%}&93\%&99\%& 96\%& 95\% \\
          \hline
         min size CNF &1744/4944& 1941/5452& 2196/6102& 1978/5534& 1837/5178 \\
    
      \hline
         avg size CNF &1979/5540& 2172/6050& 2481/6856 & 2270/6293 &2059/5727 \\
          \hline
          max size CNF&2176/6066& 2429/6760& 2789/7694& 2558/7028& 2330/6408\\
          \hline
          min enc\_runtime (s)&0.83& 0.88& 0.92& 0.82& 0.74\\
          \hline
          avg enc\_runtime (s)&1.05& 1.06& 1.11& 0.92&0.86 \\
          \hline
          max enc\_runtime (s)&1.51& 1.92& 1.56&1.31& 1.32 \\
          \hline
          \hline
             min $\#$CFs &10&13&10&15&6 \\
          \hline
         avg $\#$CFs &35790&63916&99174&79520&4846 \\
          \hline
          max $\#$CFs &285219&546005&633416&640868&65554\\
          \hline
          min enumtime (s)&0.005&0.11&0.006& 0.11&0.008 \\
          \hline
          avg enumtime (s)&21.49&42.11&77.72&50.86&2.35 \\
          \hline
          max enumtime (s)&234.18&600&600&531.16&35.08  \\
          \hline
       \end{tabular}}
       \label{table3}
        \caption{Evaluating (1) the encoding into the logical representation and (2) the enumeration of explanations for different classifiers used to locally explain MNIST images.
        }
\end{center}
\end{table}

\noindent {\bf Evaluating the feasibility of the enumeration of explanations :} 
We want to assess the practical feasibility of the enumeration of \textit{Sufficient Reasons} and \textit{Counterfactual} explanations. 
To enumerate the {$CFx$}, we use the \textit{ EnumELSRMRCache tool}\footnotemark[1] with a timeout set to 600s. \footnotetext[1]{implementing the boosting algorithm for MCSes enumeration proposed in \cite{gregoire2018boosting}} Thanks to the duality between MUSes and MCSes, the enumeration of $SRx$ can be done by computing the minimal hitting set of $CFx$. However, the results in this paper only cover the enumeration of $CFx$ due to the page limitation.
In \textbf{Table 1}, we report the average run-time (enumtime) needed to enumerate all the explanations within the timeout whereas in reality the solver manages to find explanations instantly in the majority of cases. Accordingly, the enumeration time remains reasonable and shows the practical feasibility of the enumeration of such explanations for medium size classifiers (like the BNNs used).
Additionally, we notice that the number of $CFx$ enumerated is significant and that it quickly becomes unmanageable for a user to process the result. This reinforces the need for quality metrics to filter the generated explanations.\\

\noindent {\bf{SR$_x$} and \textbf{CF$_x$} for MNIST} : 
We use the "one-vs-all" BNN $f_8$ trained to recognize the eight "8" digit (positive prediction for an image representing "8", negative otherwise) that has achieved an accuracy of $97\%$. 
Fig.\ref{xpimgs} shows a few data samples negatively predicted by $f_8$. 
Heatmaps in the $3^{rd}$ column of Fig.\ref{xpimgs} show examples of $CFx$ highlighting the necessary changes to be made on the input data sample in order to alter the outcome of $f_8$ from negative to positive. 
We can visually distinguish a sort of pattern of the digit "8" highlighting the pixels we need to act on. This actually matches the definition of $CFx$.
Although the underlying mechanisms of our approach and SHAP differ which may lead to very different explanations for the same input,  we can see that our explanations are visually simpler, clearer and easier to understand and use compared to SHAP explanations in the last column.
\begin{figure}[h!]
    \centering
    \includegraphics[width=0.95\textwidth]{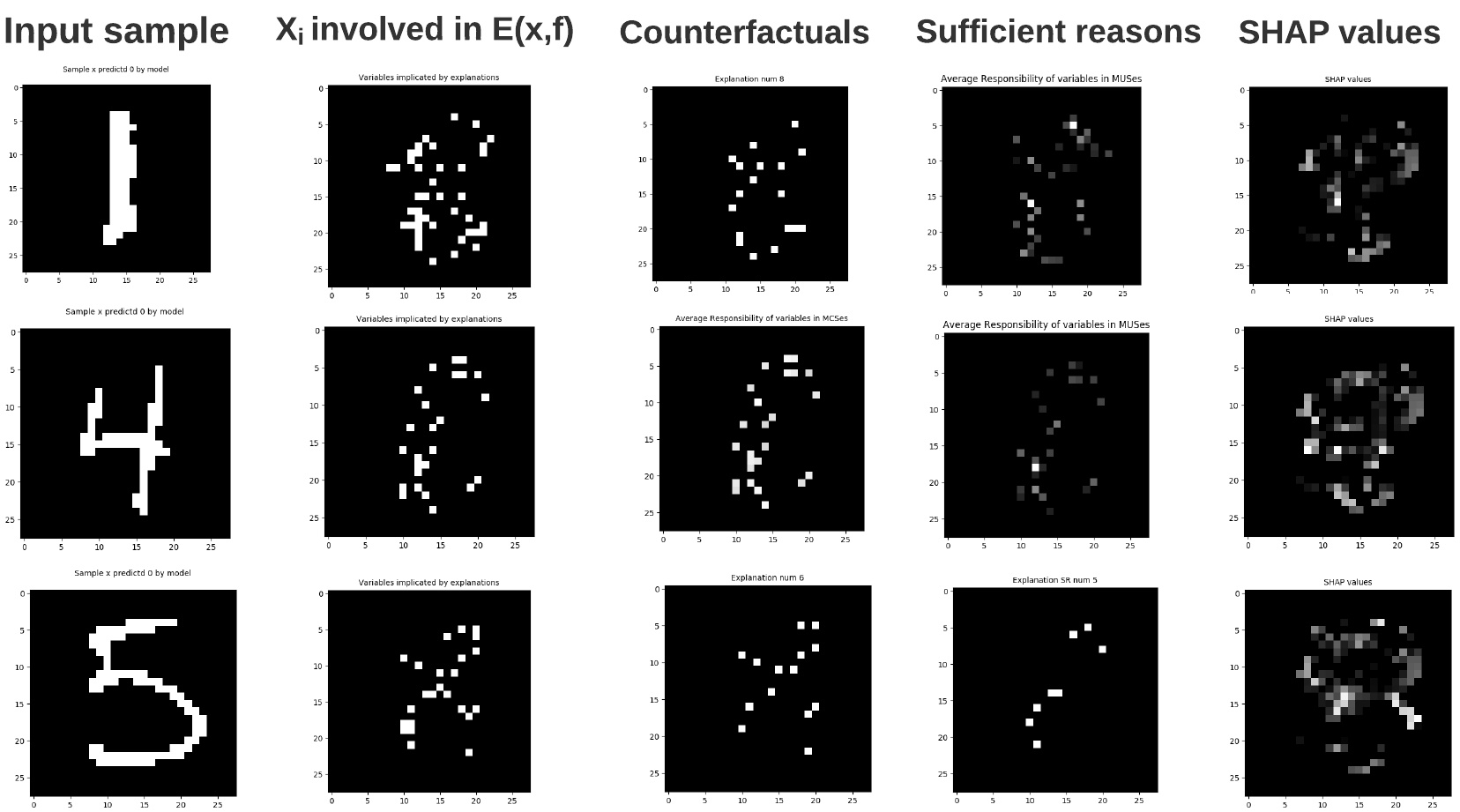}
    \caption{Data samples from MNIST predicted negatively by $f_8$ in the $1^{st}$ column. The heatmap of the: $2^{nd}$ column represent the variables involved by the explanations,  the $3^{th}$ and $ 4^{th}$ columns, a single counterfactual and sufficient reason explanation. The last column is the SHAP values of the variables contributing positively to the prediction.}
    \label{xpimgs}
\end{figure}

\section{Concluding remarks and Discussions}
We try to explain individual outcomes of black-box models by the mean of a novel model agnostic generic approach presented within this paper in order to provide two complementary types of explanations: \textit{Sufficient reasons} and \textit{Counterfactuals}. The approach is based on the Boolean satisfiability concepts which allow us to take advantage of the strengths of already existing and proven solutions, and the powerful practical tools for the generation of  MCS/MUS.
We use the notion of surrogate model to overcome the complexity of encoding a ML classifier into an equivalent logical representation. It is a local encoding since we approximate the original model in the vicinity of the sample of interest.  
The same mechanism is used to explain positively predicted instances. It suffices to work with the negation of the representation of $f$ ($\neg f$) to enumerate the explanations in a similar way.
We intend in future works to assess the relevance of explanations and features individually w.r.t a set of properties allowing 
to evaluate explanations in
ways that are closer to how users consume them. \\

\noindent{\bf Acknowledgment :} This work was supported by the Région Hauts-de-France.
%
%
%
\bibliographystyle{splncs04}
\bibliography{main}

\end{document}